\documentclass[12pt]{article}
\usepackage{graphicx}
\usepackage{url}
\usepackage{caption}
\usepackage{subcaption}
\usepackage{amsmath}
\usepackage{array}
\usepackage{verbatim}
\usepackage{authblk}
\usepackage{hyperref}

\newcommand{\blind}{0}

\begin{document}

\def\spacingset#1{\renewcommand{\baselinestretch}
{#1}\small\normalsize} \spacingset{1}

\title{\bf Pen Spinning Hand Movement Analysis Using MediaPipe Hands}

\author
{Tung-Lin “Dove” Wu$^{1}$ \qquad Taishi “Wabi” Senda$^{2}$\\
\normalsize{$^{1}$Taiwanese Pen Spinning Forum (TWPS)}\\
\normalsize{$^{2}$JapEn Board (JEB)}
\\
\medskip
\normalsize{Emails: \href{mailto:howard0100000@gmail.com}{howard0100000@gmail.com}, \href{mailto:wabi.penspinning@gmail.com}{wabi.penspinning@gmail.com}}
}
\date{}
\maketitle

\bigskip
\begin{abstract}
We challenged to get data about hand movement in pen spinning using MediaPipe Hands~\cite{zhang2020mediapipe} and OpenCV~\cite{bradski2008learning}.  The purpose is to create a system that can be used to objectively evaluate the performance of pen spinning competitions. Evaluation of execution, smoothness, and control in competitions are quite difficult and often with subjectivity. Therefore, we aimed to fully automate the process by using objective numerical values for evaluation.

Uncertainty still exists in MediaPipe's skeletal recognition, and it tends to be more difficult to recognize in brightly colored backgrounds. However, we could improve the recognition accuracy by changing the saturation and brightness in the program. Furthermore, automatic detection and adjustment of brightness is now possible. 

As the next step to systematize the evaluation of pen spinning using objective numerical values, we adopted “hand movements". We were able to visualize the ups and downs of the hand movements by calculating the standard deviation and L2 norm of the hand's coordinates in each frame. The results of hand movements are quite accurate, and we feel that it is a big step toward our goal. In the future, we would like to make great efforts to fully automate the grading of pen spinning.

\end{abstract}

\vfill

\newpage
\spacingset{1.2} 
\section{Introduction}
\label{sec:intro}
Execution and control play a big role in pen spinning, and it is often an important thing to examine whether a pen spinner could control the pen well or not by analyzing hand movement of the pen spinning video. However, pen spinning community currently does not have an objective way to tell how big the spinner's hand movement is. We want to resolve this problem by evaluating the hand movement of pen spinning videos without any subjectivity and bias, so we need to get the coordinate of the hand of each frame in the video and analyze this data. The analyzing results should not be affected by the camera angle and the distance between camera and hand.

\begin{figure}[h]
\begin{center}
\includegraphics[width=4.5in]{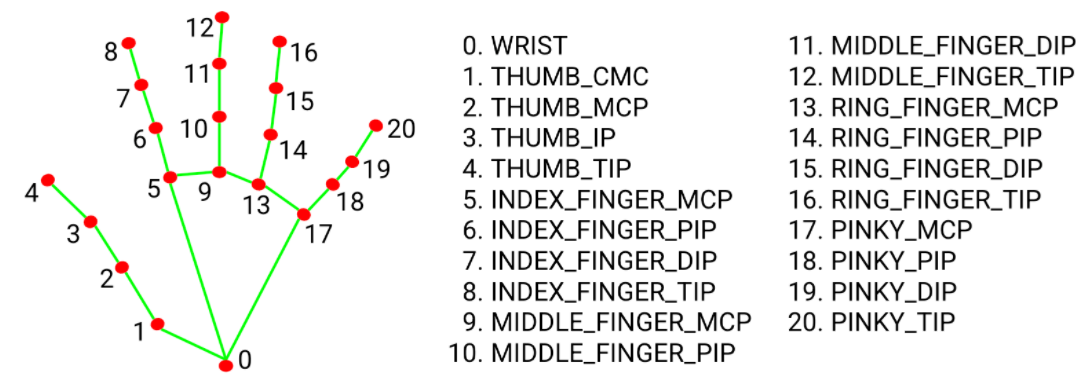}
\end{center}
\caption{Hand landmarks of MediaPipe Hands} \label{landmarks}
\end{figure}

\begin{figure}[h]
\begin{center}
\includegraphics[width=3in]{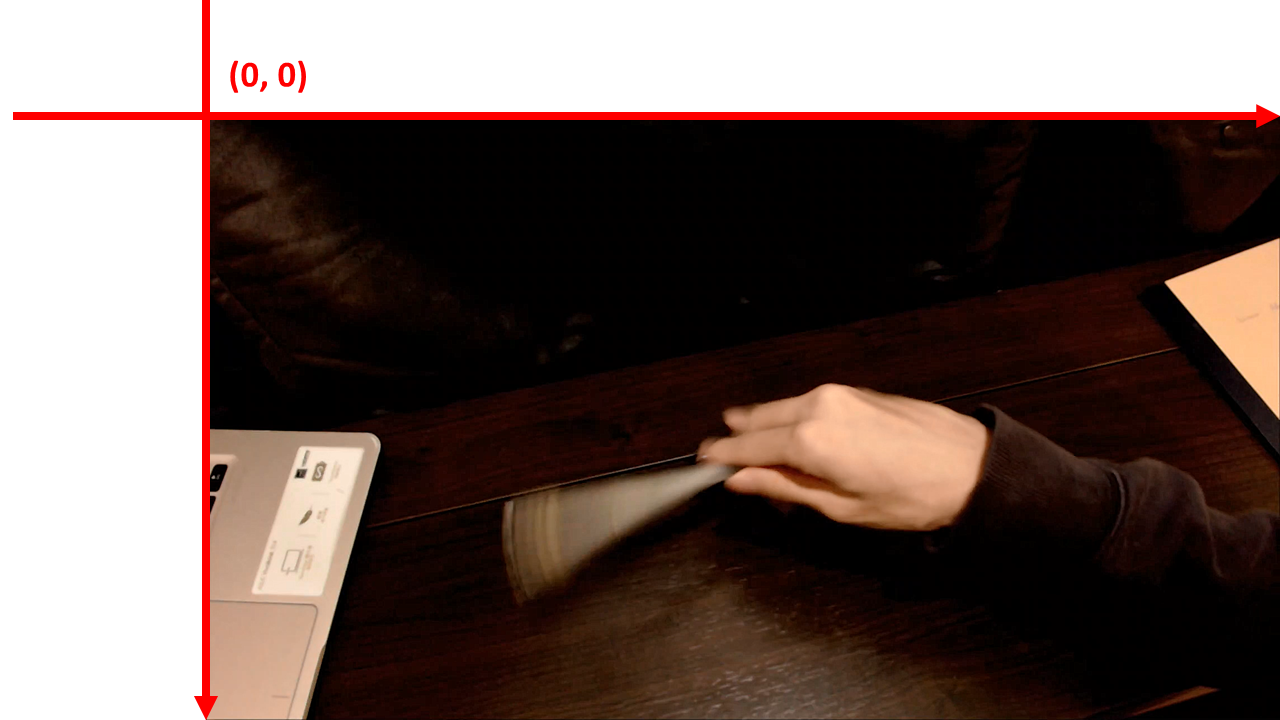}
\end{center}
\caption{Coordinate system~\cite{dovepso20r1}} \label{coordinate}
\end{figure}

We first used MediaPipe Hands~\cite{zhang2020mediapipe} made by Google to get the coordinates of the hand of the video. Details of the hand landmarks are in figure \ref{landmarks}. The origin of the vector space is the top left corner of the frame (see figure \ref{coordinate}). The coordinate of each landmark is the pixel it located. Thus the coordinates of two same video would not be the same if their resolutions are different. If we zoom in/out the video or rotate the video, the coordinate of the data would also change. In section \ref{sec:problems}, we would show how to resolve the scaling problem by normalizing the data, and how to deal with the rotation problems. Then in section \ref{sec:methods}, we designed two main methods to calculate the final result: L2 norm and standard deviation.

\section{Data Preprocessing}
\label{sec:prep}
Cutting out the parts other than pen spinning and adjusting the brightness and saturation of the video are necessary before analyzing the video. The data would have a lot of noise if we did not consider these problems.

The reason why we need to get rid of the parts that are not associative with pen spinning first is because that our research is based on skeletal estimation. If we involved coordinates of non-spinning parts, the results would be inaccurate. 

\begin{figure}[h]
    \centering
    \begin{subfigure}[30fps 720p]{0.4\textwidth}
        \centering
        \includegraphics[width=\textwidth]{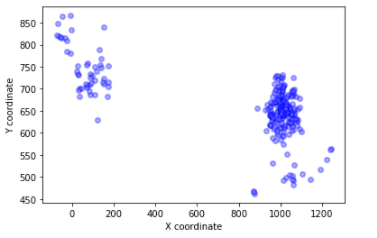}
        \caption{before : Graph of hand movements}
        \label{fuk before.png}
    \end{subfigure}
    \begin{subfigure}[30fps 720p]{0.4\textwidth}
        \centering
        \includegraphics[width=\textwidth]{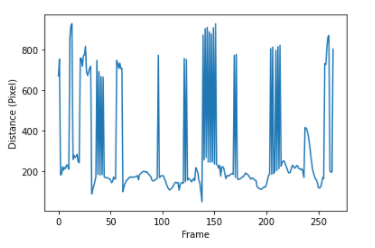}
        \caption{before : average hand position}
        \label{fuk before 2.png}
    \end{subfigure}
        \begin{subfigure}[30fps 1080p]{0.4\textwidth}
        \centering
        \includegraphics[width=\textwidth]{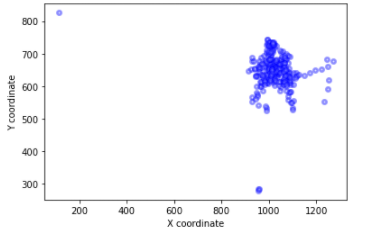}
        \caption{after : Graph of hand movements}
        \label{fuk after.png}
    \end{subfigure}
    \begin{subfigure}[rotation 1080p]{0.4\textwidth}
        \centering
        \includegraphics[width=\textwidth]{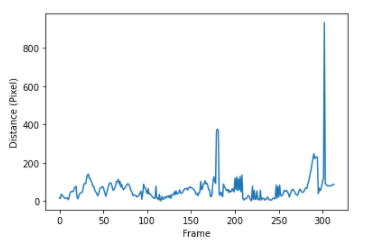}
        \caption{after : average hand position}
        \label{fuk after 2.png}
    \end{subfigure}
    \caption{fukrou for JapEn 15th~\cite{fuk15}}
\end{figure}

The next step is to change the brightness and saturation by using OpenCV~\cite{bradski2008learning}. In figure \ref{fuk before.png} and \ref{fuk before 2.png}, skeletal estimation is very unstable when the background of the pen spinning video is bright. The blue dots in the top left corner of figure \ref{fuk before.png} are all coordinates that are not belong to any part of the hand. This situation make the graph in figure \ref{fuk before 2.png} become really noisy, hence loses its reliability. In order to improve this situation, we made it possible to adjust the brightness and saturation automatically in the code. Figure \ref{fuk after.png} and \ref{fuk after 2.png} are the result after the adjustment. We successfully removed most of the noise in figure \ref{fuk after.png}, thus figure \ref{fuk after 2.png} could reflect the real result of the original video. 

\section{Scaling and Rotation problems}
\label{sec:problems}
\begin{figure}[h]
    \centering
    \begin{subfigure}[30fps 720p]{0.4\textwidth}
        \centering
        \includegraphics[width=\textwidth]{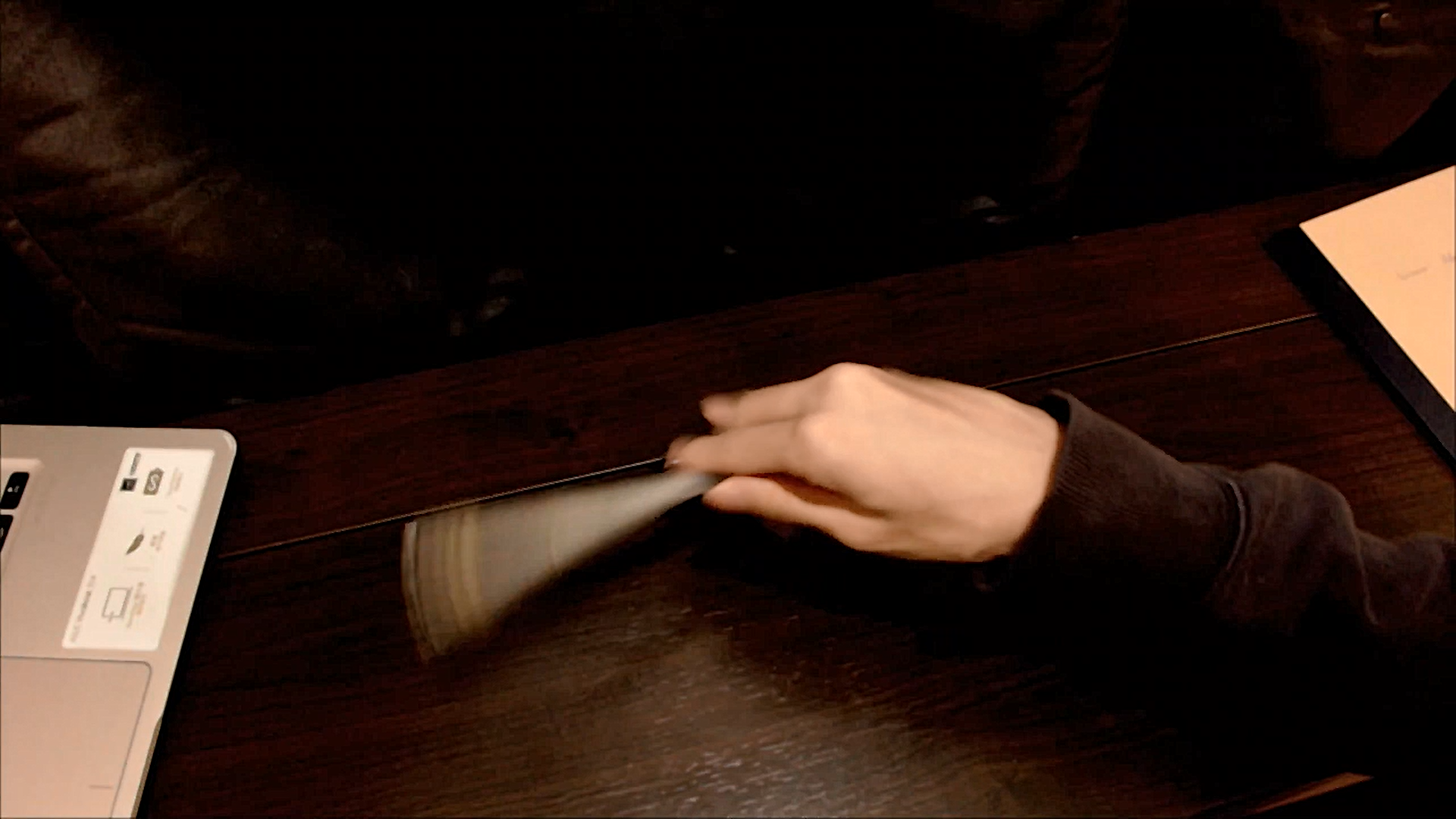}
        \caption{720p}
        \label{Dove for PSO20 R1 720p}
    \end{subfigure}
    \begin{subfigure}[30fps 1080p]{0.4\textwidth}
        \centering
        \includegraphics[width=\textwidth]{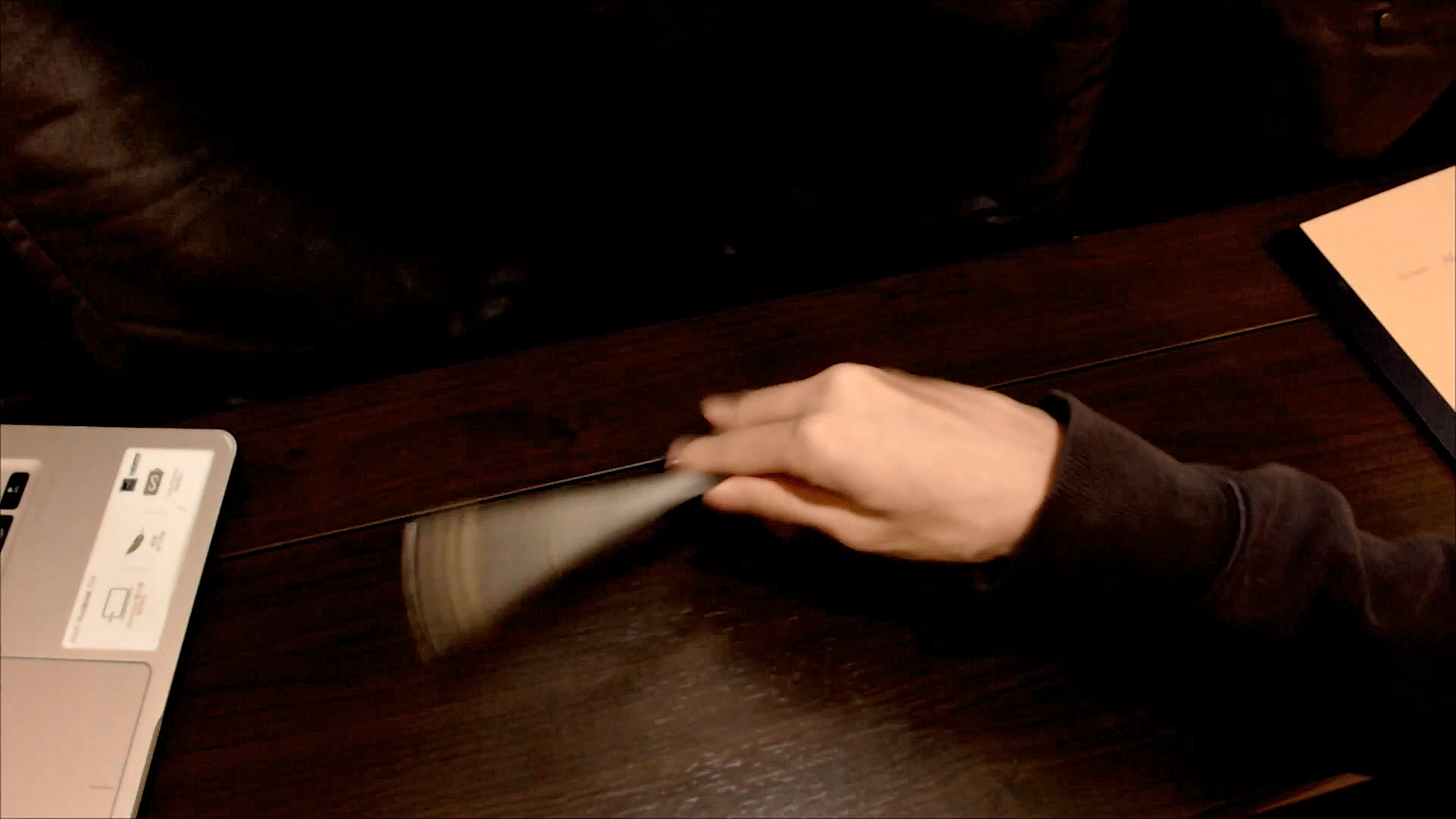}
        \caption{1080p}
        \label{Dove for PSO20 R1 1080p}
    \end{subfigure}
    \begin{subfigure}[30fps 720p]{0.4\textwidth}
        \centering
        \includegraphics[width=\textwidth]{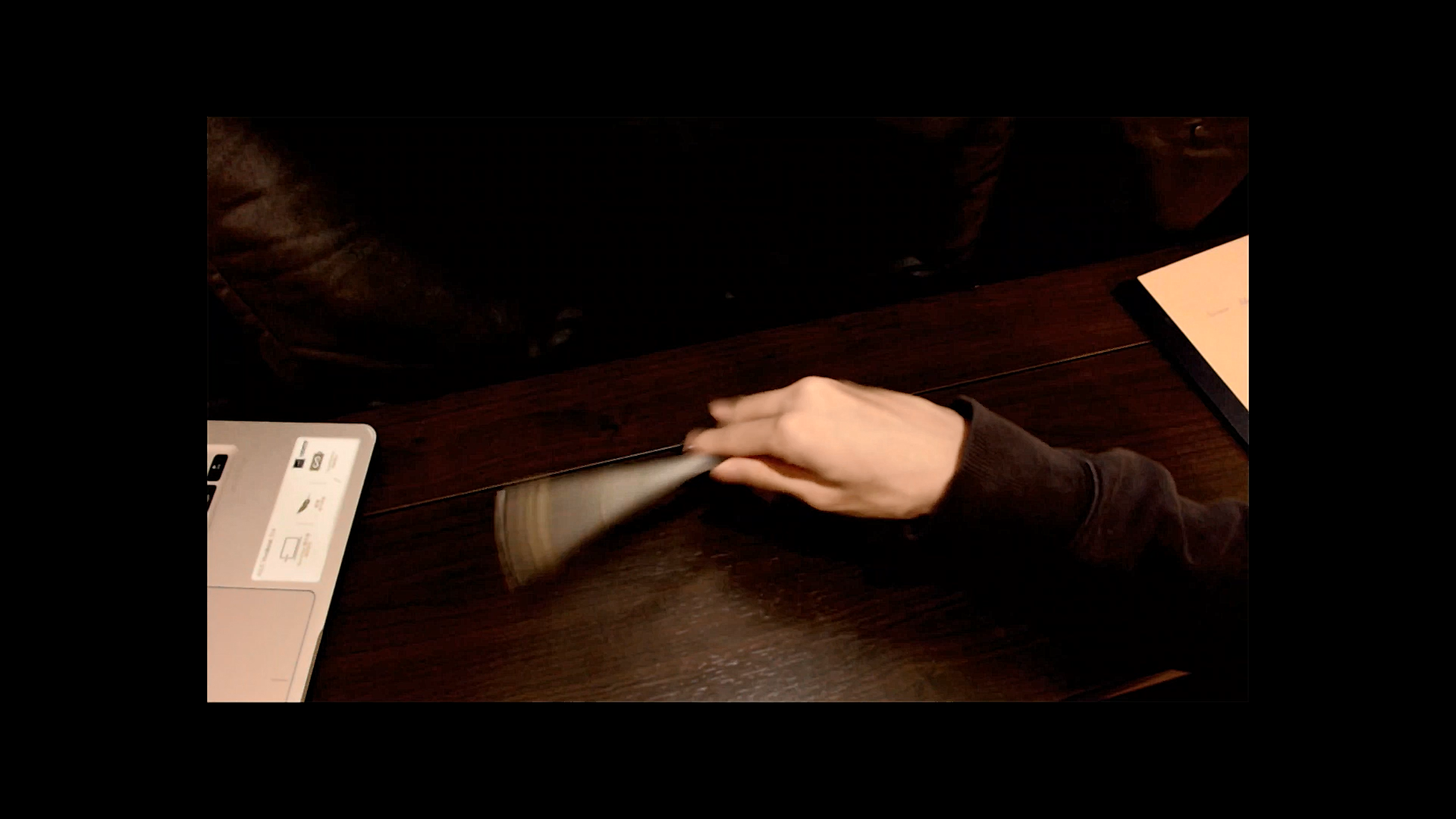}
        \caption{zoom out, 1080p}
        \label{Dove for PSO20 R1 zoom out 1080p}
    \end{subfigure}
    \begin{subfigure}[rotation 1080p]{0.4\textwidth}
        \centering
        \includegraphics[width=\textwidth]{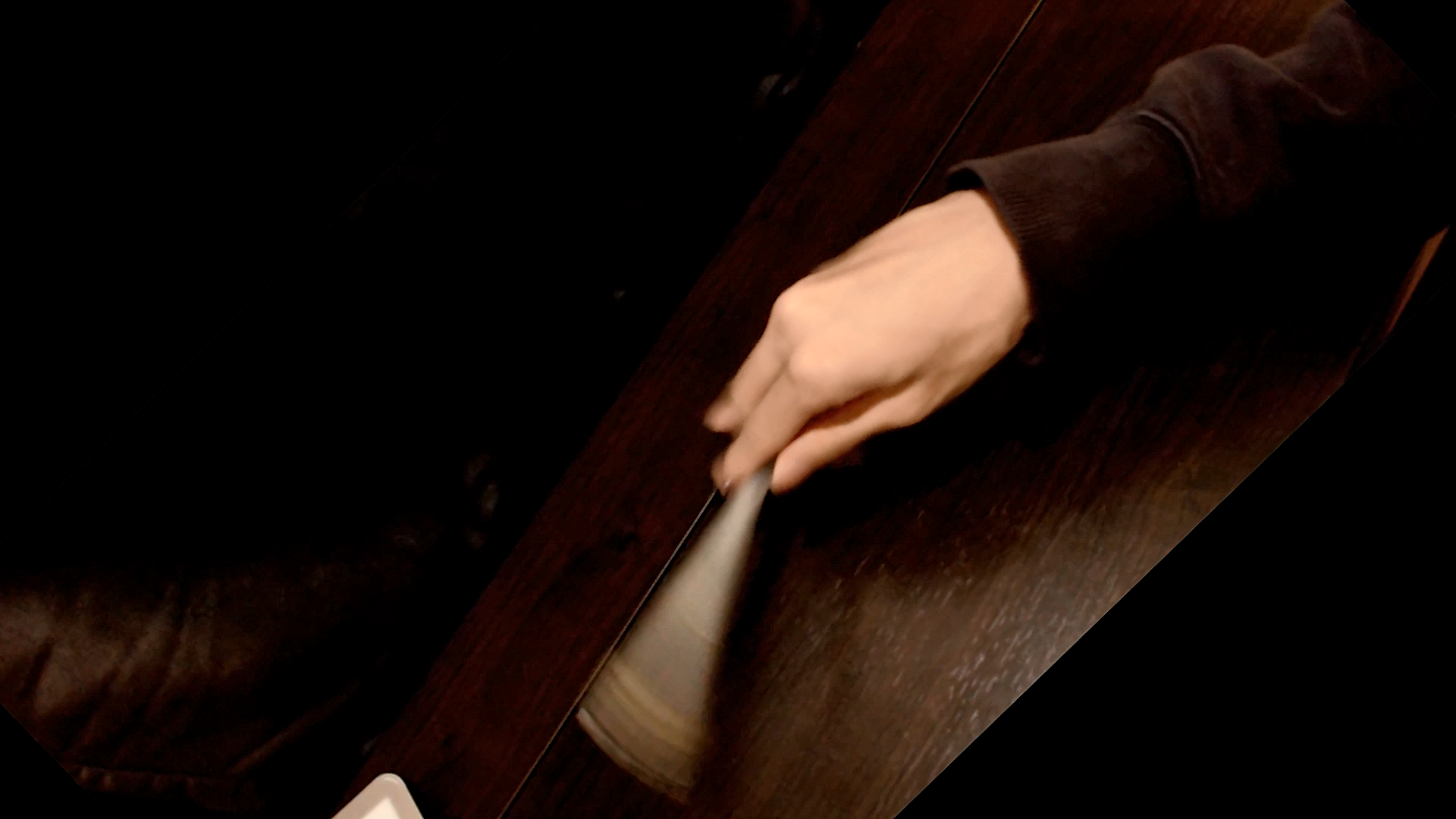}
        \caption{rotation, 1080p}
        \label{Dove for PSO20 R1 rotation 1080p}
    \end{subfigure}
    \caption{Dove for PSO20 R1~\cite{dovepso20r1}}
\end{figure}

The same video should always have the same result even if there exists some differences like scaling and rotation. We divided the scaling problem into two parts: resolution and zoom in/out. Since the coordinate of the hand is depend on which pixel the hand is, the coordinate would be different if we did not consider the resolution and zoom in/out problems. For example, figure \ref{Dove for PSO20 R1 720p}, \ref{Dove for PSO20 R1 1080p}, and \ref{Dove for PSO20 R1 zoom out 1080p} are exactly the same video, but one is 720p, the other is 1080p, and the other is zoom out a little bit. In order to make these three videos have the same results, we decided to calculate the hand size first and use the hand size to normalize the result. The formula of how we get the hand size is shown below,
\begin{equation}
    s = \frac{1}{n}\sum_{i=1}^{n}\sqrt{(x_{i, 0}-x_{i, 5})^2+(y_{i, 0}-y_{i, 5})^2}
    \label{handsize}
\end{equation}
where $x_{i, j}, y_{i, j}$ $(i = 1,2,...,n)$ are the coordinates of ith frame of landmark j (see figure \ref{landmarks}).

The rotation problem is really important because every spinner have their unique angle setup. Our method need to make sure that the result of the original video (figure \ref{Dove for PSO20 R1 1080p}) should be the same with the rotated video (figure \ref{Dove for PSO20 R1 rotation 1080p}) if they are exactly the same combo. We will discuss the details of how we resolved this situation in section \ref{sec:methods}.

Other factors like reflection, shifting, and FPS do not affect our calculations. Thus these would not be the issues while computing the results.

\section{Methods}
\label{sec:methods}

\subsection{L2 Norm}
L2 norm is a method to calculate the distance. Since any rotation of the data does not change the distance between each coordinate, there is no need to worry about it. In this method, we used landmark 0's coordinate (see figure \ref{landmarks}) of each frame of the video to be our data. The order of calculations we did is as follows.
First, we averaged the coordinates of the data to get the mean position of the hand (see equation \ref{mean}), 
\begin{equation}
\begin{aligned}
    &\bar x = \frac{1}{n} \sum_{i=1}^{n} x_{i} \\
    &\bar y = \frac{1}{n} \sum_{i=1}^{n} y_{i}
\label{mean}
\end{aligned}
\end{equation}
where $x_{i}, y_{i}$ $(i = 1,2,...,n)$ are the coordinates of ith frame of landmark 0. Then, we derived the average distance between $(\bar x, \bar y)$ and the coordinate of landmark 0 for each frame. In order to make the derived values correspond to the scaling of the video, we need to normalized the data by the hand size $s$ of the video. The details about how to get $s$ is in equation \ref{handsize}. Hence we resolved the problem by dividing the data by $s$. The whole process is in equation {\ref{distance}}, 
\begin{equation}
\begin{aligned}
    d &= \frac{1}{n} \sum_{i=1}^{n}\sqrt{(\frac{1}{s}x_{i} - \frac{1}{s}\bar x)^2 + (\frac{1}{s}y_{i} - \frac{1}{s}\bar y)^2} \\
    &= \frac{1}{n\cdot s} \sum_{i=1}^{n}\sqrt{(x_{i} - \bar x)^2 + (y_{i} - \bar y)^2}
\label{distance}
\end{aligned}
\end{equation}
where $s$ is the hand size in equation \ref{handsize}, $x_{i}, y_{i}$ $(i = 1,2,...,n)$ are the coordinates of ith frame of landmark 0.

This method is trying to measure the distance of hand movement in each frame.
Furthermore, normalizing the results ensures that every pen spinning video is in the same standard. By comparing the results with other video, we could know that whether certain moves are large or small.

\subsection{Standard Deviation}

Calculating the standard deviation is always a good method to know the unstable rate of the data. However, it only consider one dimension each time. Since we used landmark 0's coordinate (see figure \ref{landmarks}) of each frame of the video to be our data, we need to calculate the standard deviation, $\sigma_{x}$, $\sigma_{y}$, of x axis and y axis respectively and designed a function to combine those two values. The most important thing for this function is that we need to make sure the results of figure \ref{Dove for PSO20 R1 720p}, \ref{Dove for PSO20 R1 1080p}, \ref{Dove for PSO20 R1 zoom out 1080p}, \ref{Dove for PSO20 R1 rotation 1080p} should be the same.
For the scaling problem (figure \ref{Dove for PSO20 R1 720p}, \ref{Dove for PSO20 R1 1080p}, \ref{Dove for PSO20 R1 zoom out 1080p}), we only need to divide the original coordinates by the hand size. The formulas are written below, 
\begin{equation}
\begin{aligned}
    \sigma_{x} &= \sqrt{\frac{\sum_{i=1}^{n}(\frac{1}{s}x_{i} - \frac{1}{s}\bar x)^2}{n}} = \frac{1}{s}\sqrt{\frac{\sum_{i=1}^{n}(x_{i} - \bar x)^2}{n}} \\
    \sigma_{y} &= \sqrt{\frac{\sum_{i=1}^{n}(\frac{1}{s}y_{i} - \frac{1}{s}\bar y)^2}{n}} = \frac{1}{s}\sqrt{\frac{\sum_{i=1}^{n}(y_{i} - \bar y)^2}{n}}
\end{aligned}
\end{equation}
where $s$ is the hand size in equation \ref{handsize}, $x_{i}, y_{i}$ $(i = 1,2,...,n)$ are the coordinates of landmark 0 of ith frame, $\bar x$, $\bar y$ are the mean of all $x$ and $y$ of landmark 0 in equation \ref{mean}. By doing so, the data is normalized by the hand size $s$. Hence the scaling problem is resolved.

The rotation problem is more challenging while using standard deviation as the method. The data rotates about any coordinate can be treated as rotate about the origin with some shift in x and y axis. However, shifting does not change the result of $\sigma_{x}$ and $\sigma_{y}$, thus we only need to consider the situation that rotate about the origin.

The rotation matrix is shown below, 
\begin{equation}
    \begin{bmatrix}
        x_{i}'\\y_{i}'
    \end{bmatrix}=
    \begin{bmatrix}         
        \cos{\theta}&-\sin{\theta}\\\sin{\theta}&cos{\theta}
    \end{bmatrix}
    \begin{bmatrix}
        x_{i}\\y_{i}
    \end{bmatrix}
\end{equation}
where $x_{i}', y_{i}'$ $(i = 1,2,...,n)$ are the coordinates $x_{i}, y{i}$ of landmark 0 of ith frame rotated by angle $\theta$. After the rotation, the new standard deviation $\sigma_{x'}$, $\sigma_{y'}$ become
\begin{equation}
\begin{aligned}
    \sigma_{x'} &= \frac{1}{s}\sqrt{\frac{\sum_{i=1}^{n}(x_{i}' - \bar x')^2}{n}} \\
    &= \frac{1}{s}\sqrt{\frac{\sum_{i=1}^{n}((x_{i}\cos{\theta}-y_{i}\sin{\theta}) - (\bar x\cos{\theta} - \bar y\sin{\theta}))^2}{n}} \\
    &= \frac{1}{s}\sqrt{\frac{\sum_{i=1}^{n}((x_{i}-\bar x)\cos{\theta} - (y_{i}-\bar y)\sin{\theta})^2}{n}} \\
    \sigma_{y'} &= \frac{1}{s}\sqrt{\frac{\sum_{i=1}^{n}(y_{i}' - \bar y')^2}{n}} \\
    &= \frac{1}{s}\sqrt{\frac{\sum_{i=1}^{n}((x_{i}\sin{\theta}+y_{i}\cos{\theta}) - (\bar x\sin{\theta} + \bar y\cos{\theta}))^2}{n}} \\
    &= \frac{1}{s}\sqrt{\frac{\sum_{i=1}^{n}((x_{i}-\bar x)\sin{\theta} + (y_{i}-\bar y)\cos{\theta})^2}{n}} ,
\label{sd rotation}
\end{aligned}
\end{equation}
where $s$ is the hand size in equation \ref{handsize}, $\bar x'$ and $\bar y'$ are the mean of all $x'$ and $y'$ of landmark 0 in equation \ref{mean}, . The goal is trying to find a function $f(\sigma_{x'}, \sigma_{y'})$ such that 
\begin{equation}
    f(\sigma_{x'}, \sigma_{y'}) = f(\sigma_{x}, \sigma_{y})
\label{function}
\end{equation}
After some calculations, we found out that equation \ref{function} holds if $f(\sigma_{x}, \sigma_{y}) = \sqrt{\sigma_{x}^2 + \sigma_{y}^2}$. The proof is shown below. Let $A = (x_{i}-\bar x)$, $B = (y_{i}-\bar y)$ in equation \ref{sd rotation}. 
\begin{equation}
\begin{aligned}
    \sqrt{\sigma_{x'}^2 + \sigma_{y'}^2} &= \frac{1}{s}\sqrt{\frac{\sum_{i = 1}^{n}{((A\cos{\theta}-B\sin{\theta})^2+(A\sin{\theta}+B\cos{\theta})^2)}}{n}} \\
    & = \frac{1}{s}\sqrt{\frac{\sum_{i = 1}^{n}{((A^2(\cos^2{\theta}+\sin^2{\theta})+B^2(\cos^2{\theta}+\sin^2{\theta}))}}{n}} \\
    &= \frac{1}{s}\sqrt{\frac{\sum_{i = 1}^{n}{(A^2+B^2)}}{n}} \\
    &= \frac{1}{s}\sqrt{\frac{\sum_{i=1}^{n}((x_{i} - \bar x)^2 + (y_{i} - \bar y)^2)}{n}} \\
    &= \sqrt{\sigma_{x}^2 + \sigma_{y}^2}
\label{rotation result}
\end{aligned}
\end{equation}
Hence the results of figure \ref{Dove for PSO20 R1 1080p}, \ref{Dove for PSO20 R1 rotation 1080p} would be the same by using equation \ref{rotation result}.

\section{Results}
\label{sec:results}
\begin{figure}[h]
    \centering
    \begin{subfigure}[30fps 720p]{0.4\textwidth}
        \centering
        \includegraphics[width=\textwidth]{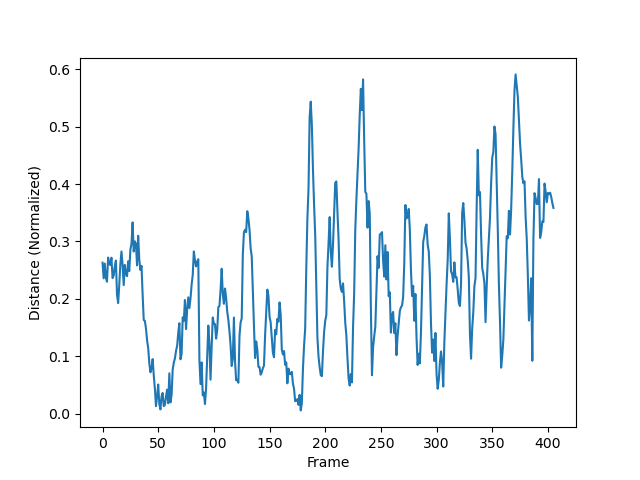}
        \caption{L2 norm}
        \label{L2 eban.png}
    \end{subfigure}
    \begin{subfigure}[30fps 720p]{0.4\textwidth}
        \centering
        \includegraphics[width=\textwidth]{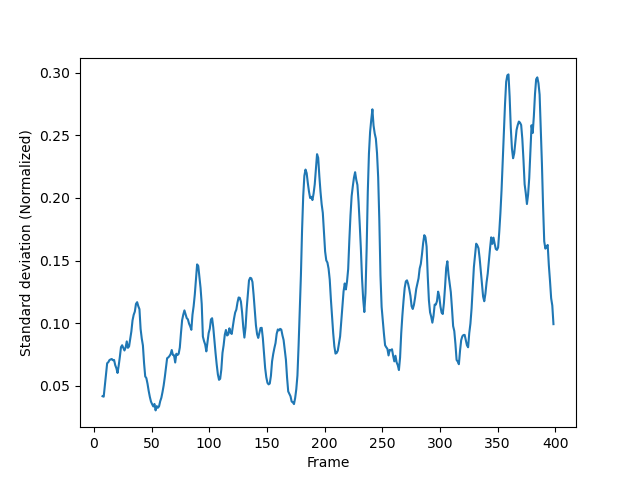}
        \caption{Standard Deviation}
        \label{SD eban.png}
    \end{subfigure}    
    \caption{eban for P.S.D.C Final~\cite{ebanpsdc}}
\end{figure}

The biggest difference between L2 norm (\ref{L2 eban.png}) and standard deviation (\ref{SD eban.png}) is that L2 norm does not consider the variation of the data. Thus using standard deviation as the metric function is more suitable in this case, which could allow us to measure the dispersion of a set of values. However, the results of these two method are really similar to each other. In figure \ref{L2 eban.png}, the y axis is the normalized hand movement distance of each frame. Figure \ref{SD eban.png} shows the standard deviation in each 15 frames. Both graphs indicate the hand movement is larger at the middle and the ending section of the video.

Table \ref{tabledove} is the result of figure \ref{Dove for PSO20 R1 720p}, \ref{Dove for PSO20 R1 1080p}, \ref{Dove for PSO20 R1 zoom out 1080p}, \ref{Dove for PSO20 R1 rotation 1080p}. The values in each row are nearly the same. The differences are come from the uncertainty of MediaPipe Hands, since some of the frames are hard to detect the hand position. However, the error is small enough so that we could ignore it without any problem.

\begin{table}[h]
\begin{center}
\begin{tabular}{c|p{2cm}<{\centering} p{2cm}<{\centering} p{2cm}<{\centering} p{2cm}<{\centering}}
 & 720p & 1080p & 1080p zoom out & 1080p rotation\\\hline
L2 Norm & 0.3732 & 0.3663 & 0.3814 & 0.3606 \\
Standard Deviation & 0.4497 & 0.4410 & 0.4479 & 0.4246 \\
\end{tabular}
\caption{Dove for PSO20 R1~\cite{dovepso20r1}}
\label{tabledove}
\end{center}
\end{table}

For table \ref{tablefukrou}, we took fukrou for JapEn 15th as an example. The result after adjusting the brightness and saturation is much more reasonable than doing nothing. We also analyzed plenty of pen spinning videos in table \ref{tableall}.
\begin{table}[h]
\begin{center}
\begin{tabular}{c|p{4cm}<{\centering} p{4cm}<{\centering}}
 & Original video & Adjust Brightness and Saturation \\\hline
L2 Norm & 1.0060 & 0.3848  \\
Standard Deviation & 1.2947 & 0.5908  \\
\end{tabular}
\caption{fukrou for JapEn 15th~\cite{fuk15}}
\label{tablefukrou}
\end{center}
\end{table}

\begin{table}[h]
\begin{center}
\begin{tabular}{c|p{2.5cm}<{\centering} p{2.5cm}<{\centering} p{2.5cm}<{\centering} p{2.5cm}<{\centering}}
& Eban for P.S.D.C Final~\cite{ebanpsdc} & iroziro for JapEn 15th~\cite{iro15} & Wabi for TsumRabbit~\cite{wabitsum} & i.suk for Simple 4th~\cite{isuksimp} \\\hline
L2 Norm & 0.2183 & 0.3079 & 0.4931 & 0.8280 \\
Standard Deviation & 0.2521 & 0.3502 & 0.5854 & 0.9786 \\
\end{tabular}
\caption{Testing results}
\label{tableall}
\end{center}
\end{table}

\section{Conclusion}
\label{sec:conc}
We have successfully used MediaPipe Hands~\cite{zhang2020mediapipe} to generate data on hand movements in pen spinning. This allows us to visualize the ups and downs of the hand movements about the pen spinning performance.

This research can help pen spinning competition's judges to judge the control of the combo objectively. Everyone who want to know how much the hand movement is in his combo in order to improve his spinning skill can also use this method. As this research continues to develop, the computer would be able to evaluate pen spinning videos individually. We can expect a fair judging system in the future since subjectivity and bias are no longer involved in a program.

We only collected data of hand position. In order to automatically evaluate the execution, control, smoothness, and presentation of pen spinning, we need to get more information on various data such as circular orbit and speed change of the pen. We leave it as a future work to finish these analyses. However, making the program able to calculate the unstable rate of a pen spinning video is definitely a big first step for this community.

Furthermore, we can not only apply this research to pen spinning, but also to dance and finger skating competitions where the stability of the performance is important once we have the data of the coordinates. Certainly it would take a tremendous amount of time to achieve this goal, but we are confident that it is completely feasible. We are all looking forward to applying cutting edge technique to pen spinning and any other field one day.


\begin{thebibliography}{99}

\bibitem{zhang2020mediapipe}
Fan Zhang, Valentin Bazarevsky, Andrey Vakunov, Andrei Tkachenka, George Sung,
  Chuo-Ling Chang, and Matthias Grundmann.
\newblock Mediapipe hands: On-device real-time hand tracking.
\newblock {\em arXiv preprint arXiv:2006.10214}, 2020.

\bibitem{bradski2008learning}
Gary Bradski and Adrian Kaehler.
\newblock {\em Learning OpenCV: Computer vision with the OpenCV library}.
\newblock " O'Reilly Media, Inc.", 2008.

\bibitem{dovepso20r1}
psdove.
\newblock [{PSO}20] {A}estheticism {R}1 {D}ove.
\newblock \url{https://youtu.be/y3ENwIk68j8}, 2020.

\bibitem{fuk15}
fukrou ps.
\newblock {J}ap{E}n15th \_ fukrou.
\newblock \url{https://youtu.be/kI5oiAP7VaU}, 2019.

\bibitem{ebanpsdc}
eban.
\newblock {P.S.D.C} vs malimo.
\newblock \url{https://youtu.be/ir6yHVJ2cec}, 2014.

\bibitem{iro15}
duriziro.
\newblock japen 15th.
\newblock \url{https://youtu.be/ovS4Dgcrba8}, 2019.

\bibitem{wabitsum}
{W}abi {PS}.
\newblock for tsumrabbit.
\newblock \url{https://youtu.be/FnjH4HEq_R8}, 2020.

\bibitem{isuksimp}
i{S}u{K}ps.
\newblock for {S}imple 4th.
\newblock \url{https://youtu.be/oWZxJvJm5Jc}, 2020.

\end{thebibliography}
\end{document}